\documentclass[10pt,twocolumn,letterpaper]{article}

\usepackage{cvpr}              

\usepackage{times}
\usepackage{epsfig}
\usepackage{graphicx}
\usepackage{amsmath}
\usepackage{amssymb}
\usepackage{booktabs}
\usepackage{booktabs,subcaption}
\usepackage{comment}
\usepackage{xcolor}
\usepackage{soul}
\usepackage{algorithm}
\usepackage{algpseudocode}

\newcommand{\method}{\texttt{ICON$^2$}\xspace}
\newcommand{\li}{\text{low-income}\xspace}
\newcommand{\mi}{\text{middle-income}\xspace}
\newcommand{\hi}{\text{high-income}\xspace}

\newcommand{\Da}{E \xspace}

\newcommand{\da}{e\xspace}
\newcommand{\interestattr}{A\xspace}
\newcommand{\interestvalue}{a\xspace}

\usepackage[hyphens]{url}

\usepackage[bookmarks=false]{hyperref}

\usepackage[capitalize]{cleveref}
\crefname{section}{Sec.}{Secs.}
\Crefname{section}{Section}{Sections}
\Crefname{table}{Table}{Tables}
\crefname{table}{Tab.}{Tabs.}

\def\cvprPaperID{1} 
\def\confName{CVPR}
\def\confYear{2023}

\begin{document}



\title{\method: Reliably Benchmarking Predictive Inequity in Object Detection \\ by Identifying and Controlling for Confounders}

\author{Sruthi Sudhakar\thanks{Corresponding author. Work done at Georgia Tech.}\\
Columbia University\\
{\tt\small sruthi@cs.columbia.edu}
\hspace{-10pt} \and 
Viraj Prabhu\\
Georgia Tech\\
{\tt\small virajp@gatech.edu}
\hspace{-10pt} \and 
Olga Russakovsky\\
Princeton University\\
{\tt\small olgarus@cs.princeton.edu}
\hspace{-10pt} \and 
Judy Hoffman\\
Georgia Tech\\
{\tt\small judy@gatech.edu}
}
\maketitle
\vspace{-5pt}
\begin{abstract}
As computer vision systems are being increasingly deployed at scale in high-stakes applications like autonomous driving, concerns about social bias in these systems are rising. 
Analysis of fairness in real-world vision systems, such as object detection in driving scenes, has been limited to observing predictive inequity across attributes such as pedestrian skin tone~\cite{Wilson2019PredictiveII}, and lacks a consistent methodology to disentangle the role of confounding variables \emph{e.g.} does my model perform worse for a certain skin tone, or are such scenes in my dataset more challenging due to occlusion and crowds? 
In this work, we introduce \method, a framework for robustly answering this question. \method leverages prior knowledge on the deficiencies of object detection systems to \emph{identify} performance discrepancies across sub-populations, compute \emph{correlations} between these potential confounders and a given sensitive attribute, and \emph{control} for the most likely confounders to obtain a more reliable estimate of model bias. Using our approach, we conduct an in-depth study on the performance of object detection with respect to income from the BDD100K driving dataset, revealing useful insights. 
\end{abstract}
\vspace{-5pt}

\vspace{-10pt}
\section{Introduction}
\vspace{-5pt}

Computer vision models today are being deployed in high-stakes applications such as self-driving cars and job hiring. These vision models rely on datasets for learning which often
contain undesirable biases that are perpetuated and in some cases amplified by these models, which can result in inequitable performance for certain sub-populations \cite{gendershades,Devries2019DoesOR,menalsolikeshopping}. 
\begin{figure}
    \vspace{-10pt}
    \centering
    \begin{center}
    \includegraphics[width=0.9\linewidth]{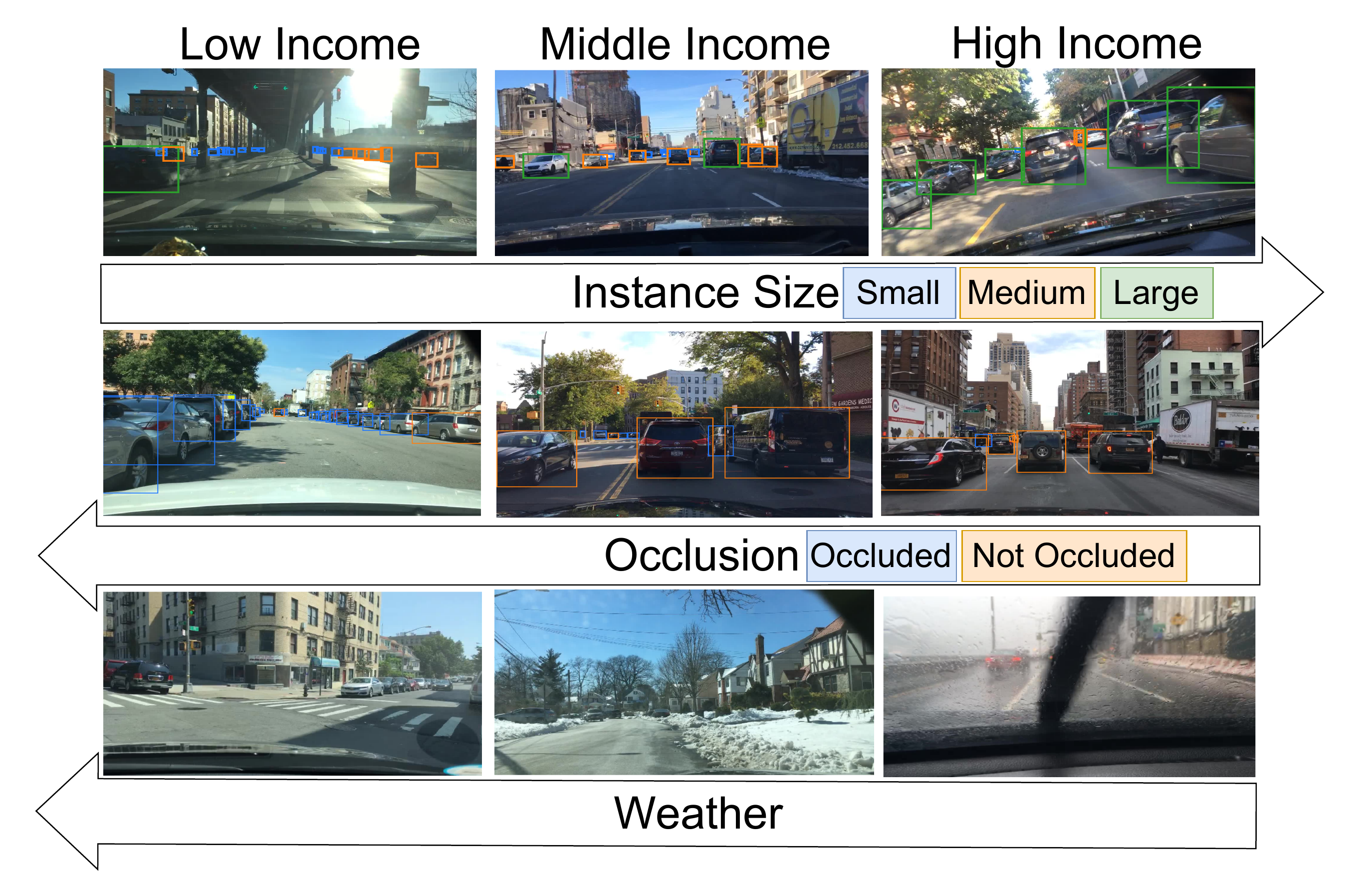}
    \end{center}
    \vspace{-10pt}
    \caption{\textbf{Overview.} We seek to reliably benchmark performance discrepancies in object detection systems used for autonomous driving across driving scenes from varying income levels.
    We propose \method, an automated framework that only uses test data and no retraining to \underline{I}dentify and \underline{Con}trol for \underline{Con}founders (such as instance size, occlusion, or weather) while evaluating predictive inequity, yielding more reliable insights.  }
    \label{fig:carvariety}
\end{figure}
Within computer vision, fairness assessment has been limited to the relatively simple task of image classification~\cite{mbgaec,tadet,towards}, with limited work in evaluating fairness in more complex tasks such as object deteciton~\cite{Wilson2019PredictiveII} 
In this work, we introduce a methodology for inspecting biases in \emph{object detection models} trained on driving datasets.

Identifying biases in detection systems is challenging due to complex performance metrics in combination with contextual~\cite{dontjudge} and co-occurrence biases~\cite{menalsolikeshopping} across visually diverse driving scenarios. 
Prior work~\cite{Wilson2019PredictiveII} has primarily relied on directly computing performance discrepancies across sub-populations to evaluate fairness across a sensitive attribute (in their case, skin tone). In practice however, this does not provide \emph{reliable} insights to inform mitigation solutions. For example, simply knowing that average precision of a detector is 2 points higher in high-income neighborhoods than in low-income ones may not tell us the complete story, without controlling for possible confounders: \eg were a higher percentage of images from low-income areas collected at nighttime, making detection harder? Similarly, were images form high-income areas less crowded or occluded, which made detection easier? Not controlling for these may lead to flawed conclusions that lead to misguided interventions (such as collecting more data from low-income regions, rather than addressing the underlying data collection protocol discrepancies). While other high-stakes applications such as medicine have long-emphasized the importance of considering confounders when assessing bias~\cite{skelly2012assessing}, autonomous driving research is yet to adopt this protocol. In this work, we provide a principled framework to identify and control for such confounding factors when measuring predictive inequity across a sensitive attribute.

Concretely, we leverage prior knowledge on known deficiencies of object detection systems, such as small objects or nighttime scenes~\cite{scaletiny,rcnnforsmalldetection,rainod,Narasimhan2004VisionAT}, 
and seek to \emph{quantify} the extent to which such attributes impact the observed system biases. Our algorithm, that we call Identify and Control for Confounders (\method), first \emph{ranks} a set of such attributes according to their potential impact on performance, and then \emph{validates} this ranking by  measuring the performance disparity while \emph{controlling} for the effects of top-ranked attributes. 

Using our approach, we conduct an in-depth study on the detection performance disparity across income levels on the BDD100K~\cite{bddk} dataset for autonomous driving. We consider innate object detection deficiencies as potential confounders, including object size, time of day, occlusion, and weather. Our method first identifies the most likely confounders, \eg finding that a larger proportion of small cars in images from lower-income regions likely contributes to worse performance. We then control for the likely confounders to see if it explains away the observed bias. Our analysis illuminates the complexity assessing fairness in object detection, and takes a step towards robust benchmarking that may guide more effective interventions.

\section{Related Work}
\label{sec:relwork}

\noindent\textbf{Fairness Toolkits.} Several analysis tools\cite{kyd,Wang2020REVISEAT,aif360-oct-2018,fbff,hardt2021amazon,udis} exist to discover biases in datasets and models. For example, Google's Know Your Data \cite{kyd} focuses on discovering correlations between attribute labels and metadata labels in TensorFlow image datasets. However, this work only focuses on ground truth label correlations, limiting the discovery of bias. Extending this idea, REVISE \cite{Wang2020REVISEAT} focuses on examining the dataset to discover biases along three dimensions: (1) object-based, (2) person-based, and (3) geography-based. However unlike us these works do not study complex tasks like object detection, and do not evaluate the impact of such correlations on performance.

\noindent\textbf{Fairness in Object Detection.} While much prior work in vision has investigated bias in image classification systems \cite{mubaw,laftr,dontjudge}, object detection models differ in model architecture and performance metrics, necessitating their own study. Prior work has investigated object detection datasets more generally \cite{Russakovsky_2013_ICCV,imagenet}, however to our knowledge only one prior work has touched on fairness in object detection. Wilson et al. \cite{Wilson2019PredictiveII} studied the predictive inequities between light-skinned and dark-skinned pedestrians using two SOTA two-stage detection systems. They aim to identify the role of two possible reasons for such inequity: time of day and occlusion, but propose a retraining-based strategy to control for these. We take inspiration from this work and develop a general methodology to reliably measure bias by controlling for various factors without  retraining.

\noindent\textbf{Common Deficiencies in Object Detection.}
There have been several works aimed at understanding factors that hurt object detection performance \cite{scaletiny,fragmented_occlusion,pdcs,ellod,sodflci}. We describe these factors in Sec \ref{sec:commondeficiencies} and leverage these works to perform our study of bias in object detection.

\vspace{-5pt}
\section{Approach} 
\vspace{-5pt}

\label{sec:approach}
We now introduce our approach to reliably benchmark predictive inequity in object detection for self-driving.

\subsection{Preliminaries}

A key aspect of our approach is to consider different underlying factors that can be used to describe aspects of the data. We refer to these quantities as \textit{\textbf{attributes}}, which can be any additional categorical variables with which the data is annotated. Example attributes include: time of day, object size, geolocation, camera resolution, \textit{etc}. In addition, we consider an attribute as \textit{\textbf{sensitive}} if it corresponds to quantities that have legal or ethical implications. Example sensitive attributes include: income level, gender, race, \textit{etc}. For clarity, we refer to attributes across which we are \emph{not} interested in inspecting bias, but which may still act as potential confounders, as \textbf{\textit{explanatory}} attributes.

\noindent\textbf{Fairness.} We define a model to be \textit{unbiased} or \textit{fair with respect} to any attribute if the performance of the system (measured by average precision in our case) is independent of the attribute. For an  attribute, $A$, with three values, $a,b, c$, the goal would be to have equal performance:
\begin{equation}
    AP_{a} = AP_{b} = AP_{c}
\end{equation}
This definition is a generalization of the equality of accuracies fairness metric~\cite{vermaWorkshop2018}. 

\noindent\textbf{Computing AP per Attribute Value.} For simplicity of notation, we present our approach considering a single class and note that it can be applied independently per-class. 
For a given category and sensitive attribute $\interestattr$, we compute AP for each value taken by the sensitive attribute $AP_{\interestvalue_i}$ by only considering images that have positive ground truth annotations for the class of interest \textbf{and} for the sensitive attribute value, $\interestvalue_i$. This strategy has a long history in object detection challenge datasets such as PASCAL~\cite{everingham2010pascal} and MSCOCO~\cite{lin2014coco} that use this approach to compute AP across object sizes (for values ``small'', ``medium'', and ``large''). 

However, AP is known to be highly sensitive to the number of positive samples in the evaluation set. 
Let $N_i$ denote the number of positive instances corresponding to attribute value $a_i$. Let $R_i(c)$ denote recall (fraction of instances detected with a confidence of at least $c$) of the object detection model over this subset and $F_i(c)$ the corresponding number of false positives. Standard precision for group $P_i(c)$ is:
\begin{equation}
    P_i(c) = \frac{R_i(c) \times N_i}{R_i(c) \times N_i + F_i(c)}  
\end{equation}

To \emph{meaningfully} compare AP values across sets with potentially different numbers of positive instances, we leverage the normalization strategy proposed by Hoeim~\emph{et al.}~\cite{Hoiem}, which computes \emph{normalized precision} $P_{N,i}(c)$, by  replacing $N_i$ with a constant N, where N is the mean of $N_i$ across all values of $\interestvalue_i$:

\begin{equation}
    P_{N,i}(c) = \frac{R_i(c) \times N}{R_i(c) \times N + F_i(c)}
\end{equation}

\noindent These normalized precision values are then interpolated and averaged across recall values to produce  $AP_{\interestvalue_i}$. 

\noindent\textbf{Variance in Attribute Performance.}
To quantify the impact of an attribute, $\interestattr$ on performance, we consider the \emph{variance} of performance across all attribute values, $\interestvalue_i \in \interestattr$
\begin{align}
    \mu_{\interestattr} &= \frac{1}{|\interestattr|}\sum_{\interestvalue_i \in \interestattr} AP_{\interestvalue_i}\\
    \sigma^2(AP_A) &= \frac{1}{|\interestattr|} \sum_{\interestvalue_i\in \interestattr} (AP_{\interestvalue_i} - \mu_{\interestattr})^2 
\label{eqn:stddev}
\end{align}

Attributes with high performance variance, $\sigma^2(AP_{\interestattr})$, have large differences in performance across different attribute values. A sensitive attribute with high performance variance suggests a possible model bias across that attribute. 

\subsection{Identifying and Controlling for Confounders}

We now introduce our \method, our two-stage framework that first identifies and ranks potential confounders, and then controls for them to obtain more reliable estimates of predictive inequity.

\subsubsection{Ranking Potential Confounders}
\label{sec:ranking}

\newcommand{\ccm}[1]{\textcolor{blue}{#1}}
\newcommand{\variance}{\text{\textsc{var}}}
\begin{algorithm}[t]
\caption{Explanatory Attribute Ranking Algorithm}
\begin{algorithmic}
\Require Explanatory Attribute Set $\mathcal{\Da}$
\Require Sensitive Attribute ($\interestattr$)
\State $\variance = [\;]$
\Comment{\ccm{variance of explanatory attribute ProxyAPs}}
\For{$\da_j \in \mathcal{\Da}$}
\Comment{\ccm{explanatory attribute: \textit{e.g., size}}}
    \State \ccm{ $\triangleright$ AP per explanatory attribute value}
    \State $\text{Compute } AP_{\da_j} \quad \forall \da_j \in \Da$     
    \State \ccm{ $\triangleright$ Empirical Distribution}
    \State \text{Compute } $P(\da_j | \interestvalue_i) \quad \forall \da_j \in \Da, \forall \interestvalue_i \in \interestattr$
    \State \ccm{ $\triangleright$ ProxyAP per sensitive value}
    \State $\text{ProxyAP}_{\interestvalue_i}^{\Da} = \sum_{\da_j \in \Da} P(\da_j|\interestvalue_i) \cdot AP_{\da_j} $ 
    \State $\mu_{\interestattr}^{\Da} = \frac{1}{|\interestattr|} \sum_{\interestvalue_i \in \interestattr} \text{ProxyAP}_{\interestvalue_i}^{\Da}$
    \State $\sigma^2(\text{ProxyAP}^{\Da}_{\interestattr}) = \frac{1}{|\interestattr|}\sum_{\interestvalue_i \in A}(\text{ProxyAP}_{\interestvalue_i}^{\Da} - \mu^E_{\interestattr}))^2$ 
    \State $\variance = [\variance,\; \sigma^2(\text{ProxyAP}^{\Da}_{\interestattr})]$ 
\EndFor \\
\Return sorted(\variance) \ccm{ $\triangleright$ Rank explanatory attributes}
\end{algorithmic}
\label{algo:1}
\end{algorithm}

Given a sensitive attribute, $\interestattr$ with non-zero attribute variance $\sigma^2(AP_{\interestattr})$ (see Eq.~\ref{eqn:stddev}) we aim to provide 
a potential explanation. To do so, our approach first ranks a set of explanatory attributes by their potential impact on model performance across the sensitive attribute. To compute this ranking, we define a proxy performance metric. 
While the variance of an explanatory attribute $\sigma^2(AP_{\Da})$, indicates the degree to which performance varies as a function of the explanatory attribute values, $e_i \in E$, this measure alone is insufficient to explain any observed variance for our sensitive attribute, $A$, as it doesn't account for the relationship between the two. 

Instead, we consider the conditional distribution of an explanatory attribute given the sensitive attribute value, $P(\Da | \interestvalue_i)$. Intuitively, if the explanatory attribute is independent of the sensitive attribute we would find that the conditional distribution does not differ based on the sensitive attribute value, so that $P(\Da | \interestvalue_i) = P(\Da|\interestvalue_j) \quad \forall \interestvalue_i, \interestvalue_j \in A$. 

For an explanatory attribute to impact the performance variance of a sensitive attribute, we need both for the explanatory attribute to have high performance variance and for the sensitive attribute to be dependent on the explanatory attribute. To capture these two notions, we introduce a metric that we call the \emph{Proxy AP}, which for a sensitive attribute value $\interestvalue_i$ and explanatory attribute $\Da$ is given by: 

\begin{equation}
    \text{ProxyAP}^\Da_{\interestvalue_i} = \sum_{\da_j \in \Da} P(\da_j | \interestvalue_i)\cdot AP_{\da_j}
\end{equation}

ProxyAP captures performance for a sensitive group $\interestvalue_i$ as a function of the explanatory attribute performance weighted according to the distribution of the explanatory attribute under the sensitive group. 
Next, we compute the performance variance (Eq.~\ref{eqn:stddev}) across proxyAP values, $\sigma^2(\text{ProxyAP}^\Da_{\interestattr})$, in order to quantify the influence of an explanatory attribute $E$ on model performance discrepancy across a sensitive attribute $\interestattr$.

Finally, we sort explanatory attributes in our set in decreasing order of $\sigma^2(\text{ProxyAP}^\Da_{\interestattr})$. This provides us with a list of explanatory attributes to investigate ranked by importance, which we utilize in the next step of our method. Our ranking approach is summarized in Algorithm~\ref{algo:1}.

\vspace{-5pt}
\subsubsection{Controlling for Confounders}
\label{sec:validation}

The explanatory attribute ranking algorithm helps to quickly arrive at directions of study for understanding and mitigating performance discrepancies. To obtain a reliable estimate of predictive inequity with respect to a sensitive attribute, we need to control for top-ranked explanatory attribute and see if they explanin away the high variance in sensitive attribute performance. 

To control for the explanatory attribute, we consider instances that are just part of one explanatory attribute value, $\da_j$, \textbf{and} part of a sensitive attribute value, $\interestvalue_i$. We can compute performance, AP, on this set of instances ($AP_{\interestvalue_i,\da_j}$). Next, to determine whether controlling for this explanatory attribute reduces variance, we use Eq.~\ref{eqn:stddev} to compute variance across settings of the sensitive attribute, $\sigma^2(AP_{\interestattr,\da_j})$, for a single explanatory attribute value. Finally, we take the mean of the variance across settings of the explanatory attribute, $\mu(\sigma^2(AP_{\interestattr,\da_j}) \; \forall \da_j \in \Da)$, and compare this average to the original variance across the sensitive attribute before controlling, $\sigma^2(AP_{\interestattr})$.

If the variance in performance after controlling for the explanatory attribute, $\mu(\sigma^2(AP_{\interestattr,\da_j}) \forall \da_j \in \Da)$, is less than the original variance in performance, $\sigma^2(AP_\interestattr)$, there is strong evidence that a possible reason for the measured performance discrepancy in the sensitive attribute is due to the known  
poor performance of this explanatory attribute in object detection. However, if the variance still persists, it shows that while this explanatory attribute may be correlated with the sensitive attribute, it is not substantially affecting the performance of the model.  

\subsection{Common Explanatory Attributes in Detection}
\label{sec:commondeficiencies}

Our approach relies on a set of explanatory attributes. 
We propose to leverage the vast literature on factors that are common failure modes of object detection systems, as a guide to further investigate and explain performance discrepancies across a sensitive attribute. 
We now detail a (non-exhaustive) list of factors which are known to negatively impact detection performance: Instance Size \cite{scaletiny,rcnnforsmalldetection,Li_2017_CVPR,Bai_2018_ECCV}, Occlusion \cite{fragmented_occlusion,occlusion_review,saod}, Crowded scenes \cite{pdcs,Chu_2020_CVPR,Xia_2018_CVPR}, Illumination \cite{ellod,immvp}, Weather \cite{rainod,Narasimhan2004VisionAT}, Contrast \cite{sodflci,Li_2016_CVPR}, Scene Layout \cite{bao2011toward,Wang2017EfficientSL}, Aspect Ratio \cite{Wang_2013_ICCV,odht,mkfod}. 

Many of these attribute values are implicit with object ground truth labels (\eg instance size, aspect ratio, contrast, scene illumination, occlusion, and crowds can all be exactly or approximately measured). Other quantities like the weather or time of day can be usually extracted from the meta data associated with the captured images. 

\section{Experiments}
\label{sec:exp}

\newcommand{\std}[1]{\textcolor{gray}{{\footnotesize $\pm #1$}}}
\begin{table}[t]
\setlength{\tabcolsep}{4pt}
\resizebox{\linewidth}{!}{
\centering
\begin{tabular}{ lcccccc } 
\toprule
    & & \multicolumn{3}{c}{$AP_{\interestvalue}$ ($A=$ Income)} & \\
    \cline{3-5}
Class & $AP$ & low & middle & high & $\sigma(AP_{\interestattr})$ \\
\cmidrule(l{4pt}r{4pt}){1-1}
\cmidrule(l{4pt}r{4pt}){2-2}
\cmidrule(l{4pt}r{4pt}){3-3}
\cmidrule(l{4pt}r{4pt}){4-4}
\cmidrule(l{4pt}r{4pt}){5-5}
\cmidrule(l{4pt}r{4pt}){6-6}
All Classes & 36.5 & 38.4 \std{0.1} & 37.3 \std{ 0.2 }& 35.4 \std{ 0.1} & 1.52\\ 
Car & 49.8 & 48.7 \std{ 0.1} & 49.6 \std{ 0.1} & 53.2 \std{ 0.1} & 2.38 \\ 
Pedestrian & 34.4 & 38.6 \std{ 0.2} & 37.3 \std{ 0.2} & 29.7 \std{ 0.1} & 4.80\\ 
Truck & 45.0 & 44.8 \std{ 0.3} & 45.7 \std{ 0.3} & 48.5 \std{ 0.2} & 1.90 \\ 
Traffic Sign & 37.3 & 38.0 \std{ 0.1} & 37.8 \std{ 0.1} & 35.2 \std{ 0.1} & 1.53 \\ 
Traffic Light & 25.1 & 25.6 \std{ 0.1} & 25.8 \std{ 0.1 }& 23.5 \std{ 0.1} & 1.27\\ 
Bus & 48.2 & 51.3 \std{ 0.4} & 49.3 \std{ 0.5} & 46.3 \std{ 0.4} & 2.49 \\ 
Motorcycle & 24.3 & 26.7 \std{ 0.8} & 25.5 \std{ 0.8} &  23.9 \std{ 0.6} & 1.43 \\ 
Bicycle & 25.4 & 33.5 \std{0.7} & 26.6 \std{0.8} & 23.6 \std{ 0.3} & 5.05\\
\bottomrule
\end{tabular}}
\vspace{-5pt}
\caption{ Analysis of model performance on Income attribute across all 8 classes (row 1), and on each class independently in BDD100K. For each row we measure overall AP within each of the three sensitive attributes (`low', `middle' and `high' income) and standard deviation in performance across these values.}

\label{table:apacrossincome}
\end{table}

\begin{figure*}[t]
    \centering
    \includegraphics[width=0.8\linewidth]{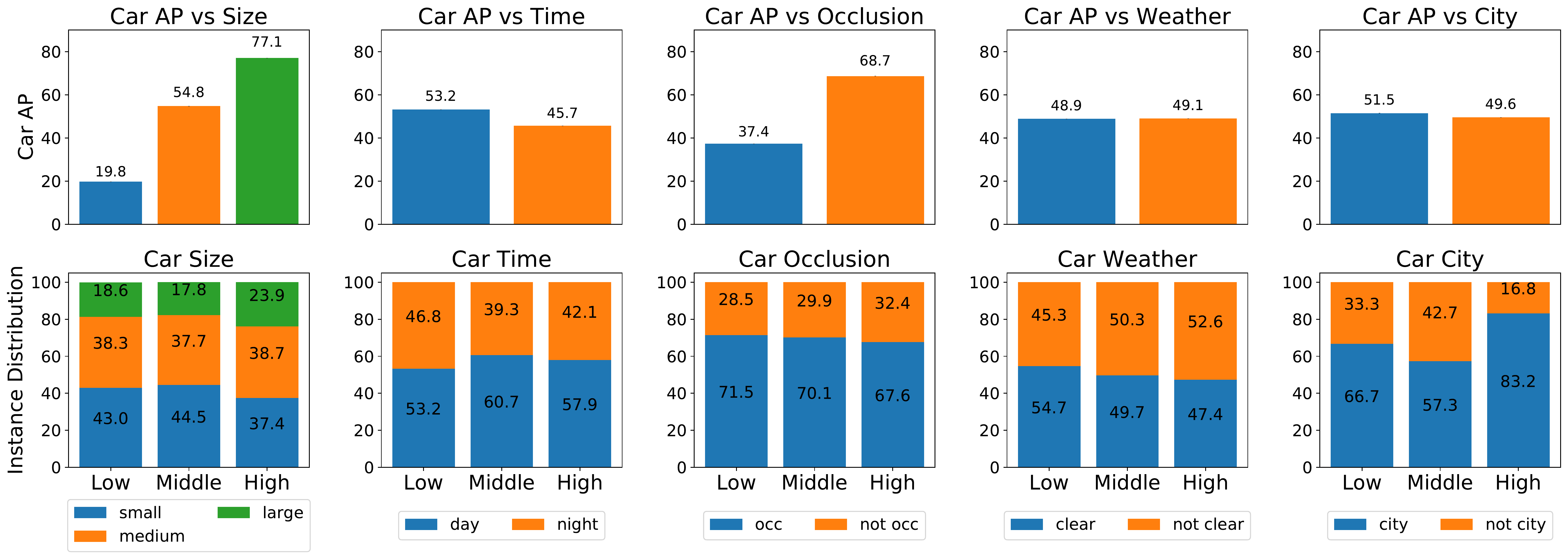}
    \caption{Analysis of the \textbf{Car} class in BDD100K.
    Performance vs explanatory attribute ($AP_{\da_i}$, \textit{top row}) and the distribution of the explanatory attribute values within each sensitive attribute value ($P(\Da |\interestvalue)$, \textit{bottom row}). 
    }
    \label{fig:car_gt_and_apda}
\end{figure*}

\begin{figure*}
\centering
    \includegraphics[width=0.8\linewidth]{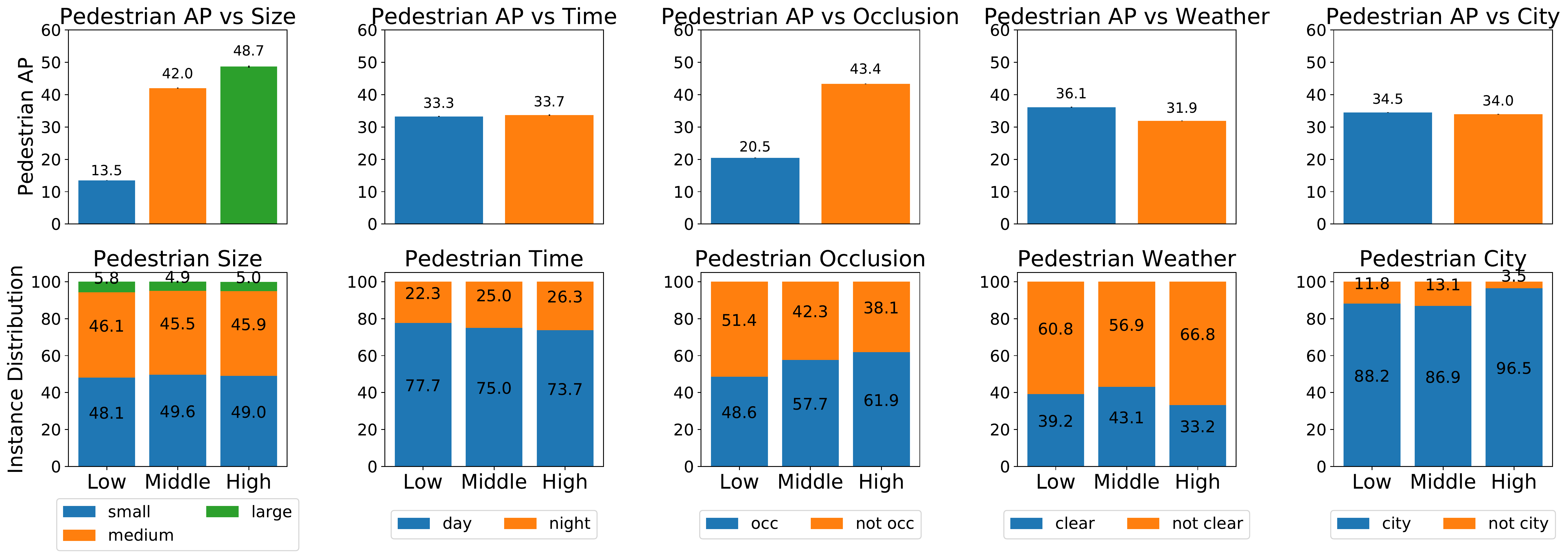}
    \caption{ Analysis of \textbf{Pedestrian} class in BDD100K.
    Performance vs explanatory attribute ($AP_{\da_i}$, \textit{top row}) and the distribution of the explanatory attribute values within each sensitive attribute value ($P(\Da |\interestvalue)$, \textit{bottom row}).
    }
    \label{fig:ped_gt_and_apda}
\end{figure*}

\begin{figure*}[h]
\centering
\vspace{-5pt}
    \includegraphics[width=0.9\linewidth]{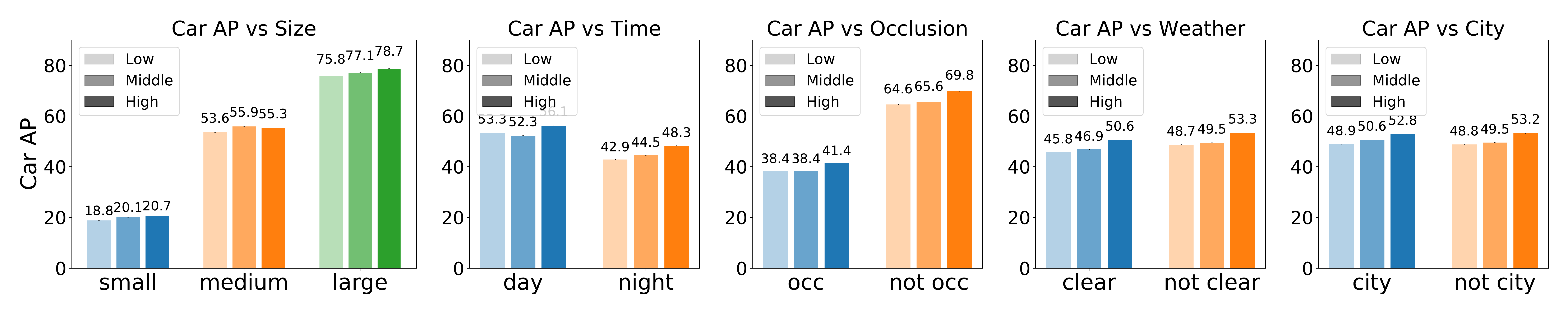}
    \caption{\textbf{Controlled AP} of the \textbf{Car} class across income levels while controlling for 5 explanatory attributes. 
    }
    \vspace{-5pt}
    \label{fig:carcontrolda}
\end{figure*}

\begin{figure*}[h]
\centering
\vspace{-5pt}
    \includegraphics[width=0.9\linewidth]{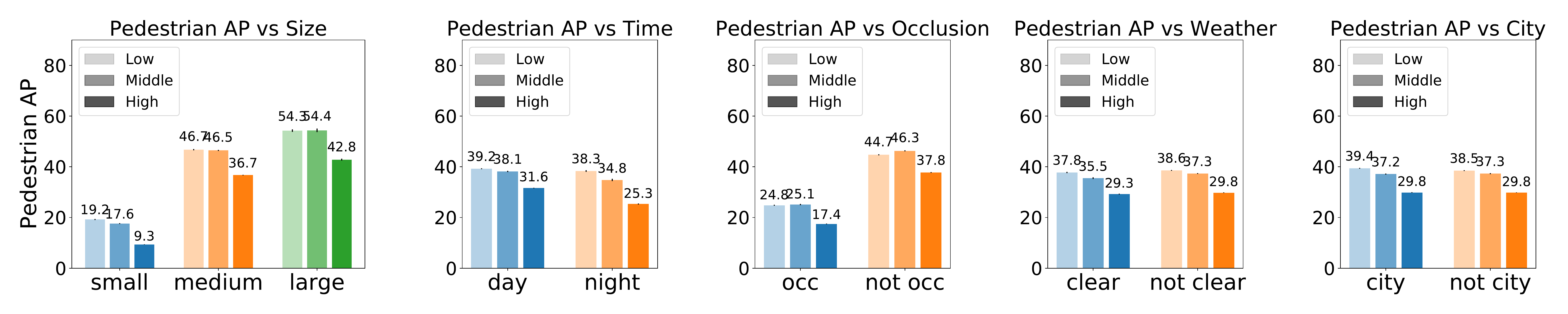}
    \vspace{-5pt}
    \caption{ \textbf{Controlled AP} of the \textbf{Pedestrian} class across income levels while controlling for 5 explanatory attributes.
    }    
    \vspace{-5pt}
    \label{fig:pedcontrolda}
\end{figure*}

\subsection{Dataset \& Implementation Details}

We evaluate our framework on the BDD100K Driving Dataset~\cite{bddk} which contains bounding box annotations for 100k images and 13 classes (`bicycle', `bus', `car', `motorcycle', `other person', `other vehicle', `pedestrian', `rider', `traffic light', `traffic sign', `trailer', `train', `truck'), from diverse scene types, weather conditions, and times of day. 

We study how object detection performance varies across the sensitive attribute of income level ($\interestattr$). 
We limit our data to New York City to normalize for cost of living. To obtain income annotations, we follow REVISE~\cite{Wang2020REVISEAT}, and correlate the GPS coordinates for each image (available for 68\% of the dataset) with a median income ranging from \$21.4k up to \$250k. After associating these images with a median income value, 
 we construct three equally-sized income bins: \li (\$21.4k-\$62.1k), \mi (\$62.1k-\$93.4k), and \hi (\$93.4k-\$250k). 

We use the Detectron2 library~\cite{detectron2} and employ with a Faster-RCNN model~\cite{fasterRCNN} pretrained on ImageNet~\cite{imagenet} with a ResNet~\cite{resnet} + FPN~\cite{fpn} backbone. We finetune this model on the 70k training set images in BDD100K with GPS annotations and evaluate on the validation set of 20k images. We limit our evaluation to 8 classes (`bicycle', `bus', `car', `motorcycle',  `pedestrian', `traffic light', `traffic sign', `truck'), removing classes with very few examples or high label noise. To verify the effectiveness of the trained detector, we compute mean average precision (mAP) across all 8 classes in the BDD100K validation set, observing an mAP of 36.5, which is on-par with modern baselines.

\vspace{-5pt}
\subsection{Results}

First, we compute both example performance per explanatory attribute ($AP_{e_j}$) as well as the distribution of each explanatory attribute by income level ($P(E | a_i$)) for all classes. A visualization of the results for the Car class (Fig.~\ref{fig:car_gt_and_apda}) and the Pedestrian class (Fig.~\ref{fig:ped_gt_and_apda}) are provided as reference. We restrict our study in the main paper to the `car', `pedestrian', and `truck' classes (rest in appendix) as they exhibit a potential bias (indicated by high variance) and have sufficient instances in each sensitive value even after controlling for the explanatory attribute, allowing for reliable AP estimates. 

Next, we compute mean AP on the validation set over the 8 classes from BDD100K at a 95\% confidence interval for each income value (`low', `middle', and `high'), and the variance in performance of these values as described in Sec. \ref{sec:approach}.  Results are presented in Table~\ref{table:apacrossincome}. As shown, we get mAP values of 38.4 \std{ 0.1}, 37.29 \std{ 0.2}, and 35.43 \std{ 0.1}  for the low, middle, and high income values respectively (Table \ref{table:apacrossincome}, row 1). Notice that as income increases, performance \emph{decreases}. This inverse relationship between performance and income suggests that the model could be performing worse on higher income groups, but does not account for underlying confounders that may explain away this discrepancy. Moreover, each of the 8 classes do not follow the same relationship (\emph{e.g.} income and performance are directly correlated for the car class). We now proceed to run our proposed method \method to discover and control for confounders in order to obtain reliable fairness estimates.

\vspace{-5pt}
\subsubsection{Ranking Confounders}
\vspace{-5pt}

To simplify analysis, we consider one class at a time. For each class, we identify and rank explanatory attributes that could explain away the observed performance gap across income levels. 
We consider five (non exhaustive) explanatory attributes (described in Sec \ref{sec:commondeficiencies}) as possible confounders, to study further (instance `size', `time of day', `occlusion', `weather', and `scene') for which we have dataset annotations. We follow the strategy described in Algorithm~\ref{algo:1} to rank potential confounders in descebding order of performance variance in ProxyAP. We report these ordered values for the car, pedestrian, and truck, in the second column of Table~\ref{table:control_results}. As seen, our preliminary ranking indicates that size is likely to be the largest confounder for the car and truck classes, whereas for pedestrian it is likely to be occlusion.
Note that large $\sigma(\text{ProxyAP}^\Da)$ (greater than 1) for the top ranked attribute indicates that $\Da$ is a more probable explanation for variance in performance across income. We hypothesize that attributes with lower rankings and a smaller $\sigma(\text{ProxyAP}^\Da)$ do not contribute substantially to the observed performance discrepancy.

\vspace{-5pt}
\subsubsection{Controlling for Confounders}
\vspace{-5pt}

\begin{table*}[h!]
    \centering
    \begin{subtable}[t]{.32\linewidth}
    \resizebox{\textwidth}{!}{
    \begin{tabular}{r|c c c c}
    \multicolumn{5}{c}{$\sigma(AP_{\interestattr}) = $ \textbf{2.38}}\\
    \toprule
     & $\Da$ & {\footnotesize $\sigma(\text{ProxyAP}^\Da)$} & $\mu(\sigma(AP_{\interestattr,\da_j}))$ & $\Delta$\\
    \midrule
    \texttt{1} &  size      & \textbf{1.65}  & \textbf{1.20} & \textbf{1.18} \\
    \texttt{2} &  occlusion & 0.51  & 2.24 & 0.14 \\
    \texttt{3} &  time      & 0.23  & 2.38 & 0.00 \\
    \texttt{4} &  scene     & 0.20  & 2.17 & 0.22 \\
    \texttt{5} &  weather   & 0.00  & 2.50 & -0.11 \\
        \bottomrule
    \end{tabular}
    }
    \caption{ Car}
    \label{tab:CarOrder}
    \end{subtable}
    \hfill
    \begin{subtable}[t]{.32\linewidth}
    \resizebox{\textwidth}{!}{
    \begin{tabular}{r|c c c c}
   \multicolumn{5}{c}{ $\sigma(AP_{\interestattr}) = $ \textbf{4.80}}\\
    \toprule
     & $\Da$ & {\footnotesize $\sigma(\text{ProxyAP}^\Da)$} & {\footnotesize $\mu(\sigma(AP_{\interestattr,\da_j}))$} & $\Delta$\\
    \midrule
    \texttt{1} &  occlusion & \textbf{1.27} & \textbf{4.45} & \textbf{0.35}\\
    \texttt{2} &  size      & 0.21 & 5.89 & -1.09 \\
    \texttt{3} &  weather   & 0.17 & 4.59 & 0.21 \\
    \texttt{4} &  scene     & 0.02 & 4.85 & -0.05 \\
    \texttt{5} &  time      & 0.01 & 5.42 & -0.62 \\
        \bottomrule
    \end{tabular}
    }
    \caption{ Pedestrian}
    \label{tab:PedOrder}
    \end{subtable}
    \hfill
    \begin{subtable}[t]{.32\linewidth}
    \resizebox{\textwidth}{!}{
    \begin{tabular}{r|c c c c}
    \multicolumn{5}{c}{ $\sigma(AP_{\interestattr}) = $ \textbf{1.90}}\\
    \toprule
     & $\Da$ & $\sigma(\text{ProxyAP}^\Da)$ & $\mu(\sigma(AP_{\interestattr,\da_j}))$ & $\Delta$\\
    \midrule
    \texttt{1} &  size      & \textbf{1.21} & \textbf{1.02} & \textbf{0.89}\\
    \texttt{2} &  scene     & 0.44 & 1.86 & 0.05 \\
    \texttt{3} &  weather   & 0.37 & 1.82 & 0.08 \\
    \texttt{4} &  occlusion & 0.34 & 2.15 & -0.25 \\
    \texttt{5} &  time      & 0.09 & 2.24 & -0.34\\
        \bottomrule
    \end{tabular}
    }
    \caption{ Truck}
    \label{tab:TruckOrder}
    \end{subtable}
    \vspace{-10pt}
    \caption{We show reduction in variance after controlling for the 5 chosen explanatory attributes on 3 classes (car, pedestrian, and truck). 
    Notice that the top ranked attribute consistently leads to the largest reduction in variance in performance. For car and truck this corresponds to the size explanatory attribute while for pedestrian it corresponds to the occlusion explanatory attribute.}
    \label{table:control_results}
\end{table*}

Next, we follow the methodology proposed in Section~\ref{sec:validation} and compute $\mu(\sigma^2(AP_{\interestattr,\da_j}))$ for the `car', `pedestrian', and `truck' classes. To do so, we compute controlled AP (results for car and pedestrian classes reported in Fig.~\ref{fig:carcontrolda} and Fig.~\ref{fig:pedcontrolda}), and summarize the reduction in variance after controlling for each confounder in Table~\ref{table:control_results}. We find: 

\noindent \textbf{Car:} The `size' explanatory attribute is ranked as the most probable (Table~\ref{tab:CarOrder}) reason for the large (2.38, Table~\ref{table:apacrossincome}) variance in performance for the car class. Notice that after controlling for size, the variance in performance reduced by 1.18 (Table \ref{tab:CarOrder}), which is about a 50\% reduction in variance! This
validates that a significant contributing factor to the previously measured predictive inequity across income levels is the fact that the object size distribution varies significantly across income regions. 

\noindent \textbf{Pedestrian:} Unlike car and truck, the top ranked attribute for the pedestrian class is `occlusion'. After controlling for occlusion, we notice that the variance reduces by 0.35, which is about $\sim$10\% (Table \ref{tab:PedOrder}). This validates that there exists some spurious correlations between the occlusion and income attribute. However, it is likely that there are other attributes that are also causing performance gaps. For example, controlling for weather (the 3rd ranked attribute) reduces variance by 0.21 (full results in appendix).

\noindent \textbf{Truck:} Similarly, the `size' explanatory attribute is ranked as the most probable reason for the 1.9 variance in performance for the truck class. We notice that variance in performance reduces by 0.89 (Table \ref{tab:TruckOrder}), which is also about a 50\% reduction in variance! Therefore, small objects are spuriously correlated with income and partially explain away the observed performance discrepancies across income levels for the truck class.

Overall, this study shows that it is important to consider common performance reducing attributes to help us understand how explanatory attributes that affect object detection systems spuriously correlate with a sensitive attribute due to dataset imbalances, and how these correlations reflect in performance discrepancies in the model. Furthermore, by controlling for certain explanatory attributes one may notice that performance discrepancies decrease, revealing possible factors that affect the sensitive attribute. This in-depth analysis can provide insight into sources of performance discrepancies and serve as a guide for understanding and leading future mitigation efforts. Note that it is possible to further the study by controlling for more than one explanatory attribute at a time, however, it is important to ensure that enough data remains in each subset to guarantee reliable AP evaluations before making comparisons.

\vspace{-5pt}
\section{Conclusion}
\vspace{-5pt}

We propose a framework to reliably benchmark bias in object detection systems used for autonomous driving. We highlight the necessity of investigating common failures in object detection systems as possible confounders, and further provide steps to understand how these common deficiencies affect performance with respect to the sensitive attribute of interest. We present an in-depth study on the performance of object detection with respect to income from the BDD100K dataset, and highlight our findings on possible reasons for measured biases. 
\newpage
{\small
\bibliographystyle{ieee_fullname}
\bibliography{egbib}
}
\appendix
\section*{Appendix}
\label{subsec:supp}

\begin{table*}[t!]
\centering
\resizebox{0.9\textwidth}{!}{
\begin{tabular}{r| cc cc cc cc cc }
    \toprule
    &  \multicolumn{2}{c}{Traffic Sign}& \multicolumn{2}{c}{Traffic Light}& \multicolumn{2}{c}{Bus}& \multicolumn{2}{c}{Motorcycle} & \multicolumn{2}{c}{Bicycle}\\
    \cmidrule(l{4pt}r{4pt}){2-3}
    \cmidrule(l{4pt}r{4pt}){4-5}
    \cmidrule(l{4pt}r{4pt}){6-7}
    \cmidrule(l{4pt}r{4pt}){8-9}
    \cmidrule(l{4pt}r{4pt}){10-11}
    \small{Rank}   & \small{$E$} & \small{$\sigma(\text{ProxyAP}^\Da)$} & \small{$E$} & \small{$\sigma(\text{ProxyAP}^\Da)$}  & \small{$E$} &\small{$\sigma(\text{ProxyAP}^\Da)$}  & \small{$E$} &\small{$\sigma(\text{ProxyAP}^\Da)$}  & \small{$E$} &\small{$\sigma(\text{ProxyAP}^\Da)$}  \\
    \midrule
    \texttt{1}     & size     & 0.214  & time      & 0.230  & size      & 0.881  & size      & 0.372  & size      & 0.715 \\
    \texttt{2}     & weather  & 0.159  & size      & 0.205  & occlusion & 0.700  & occlusion & 0.293  & weather   & 0.456 \\
    \texttt{3}     & occlusion& 0.119  & scene     & 0.061  & weather   & 0.242  & weather   & 0.138  & time      & 0.388 \\
    \texttt{4}     & time     & 0.069  & weather   & 0.060  & scene     & 0.239  & scene     & 0.032  & occlusion & 0.201 \\
    \texttt{5}     & scene    & 0.021  & occlusion & 0.068  & time      & 0.060  & time      & 0.023  & scene     & 0.041 \\
    \bottomrule
\end{tabular}
}
\caption{Explanatory attribute output ranking for the remaining 5 classes in the BDD-100K dataset. Note: all proxyAP variance values are smaller than the discovered explanatory attributes for car, pedestrian, truck (Table~\ref{table:control_results}) and hence we did not find a compelling explanation for our sensitive attribute (income), however for the bus, motorcycle, bicycle classes the counts are too low to compute reliable AP values. }
\label{table:darankings}
\end{table*}

\section{Limitations}

\begin{figure}[h]
\centering
\includegraphics[width=\linewidth]{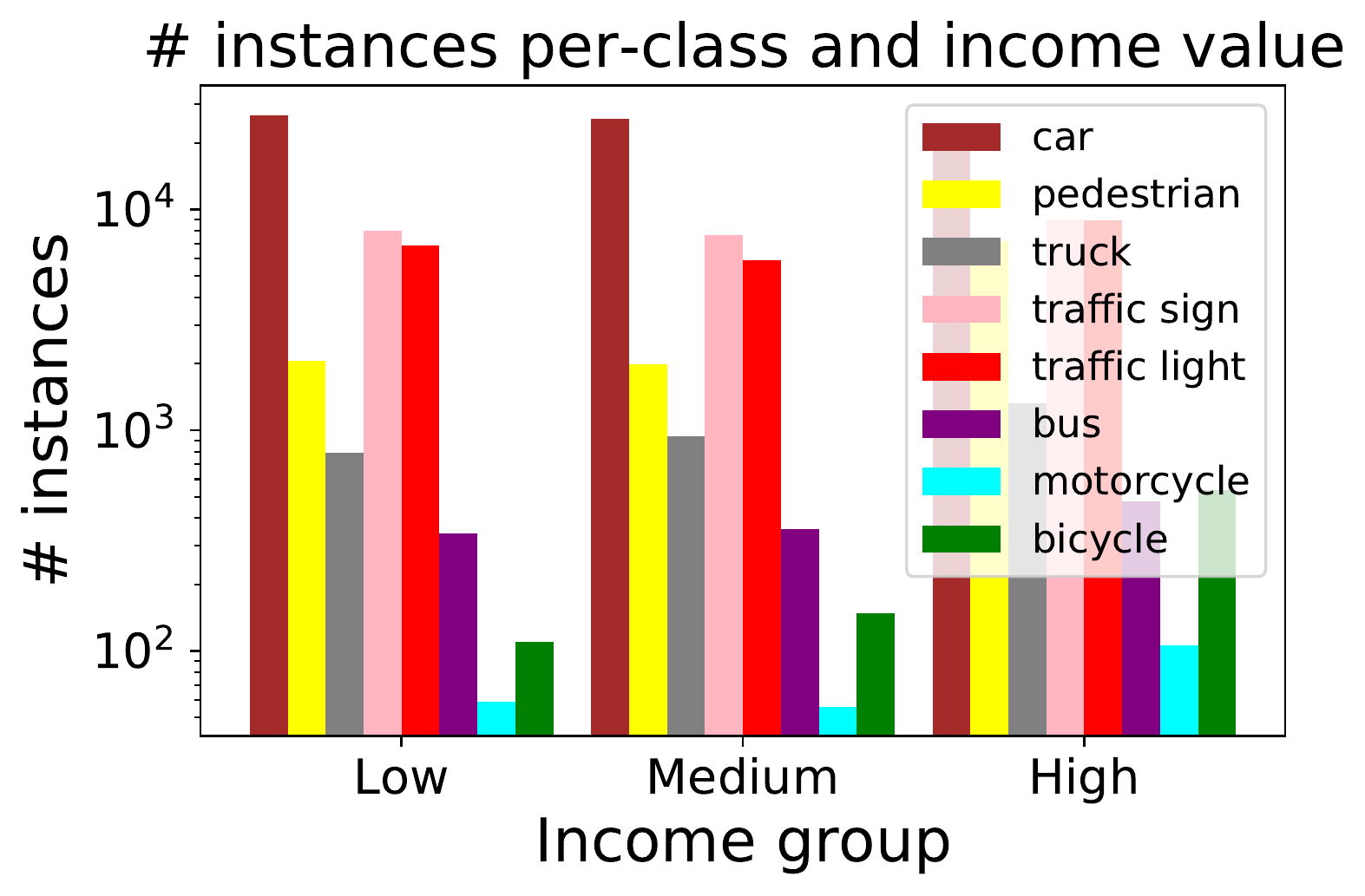}
\caption{Instance counts of each class in each income value (\emph{i.e.} low, middle, and high income). Note traffic sign, traffic light, bus, motorcycle, and bicycle have very few instances for low/middle income (there are even fewer instances when controlling for explanatory attributes). 
    }
    \label{fig:showdistributionbreakdowns}
\end{figure}

We attempt to explain measured biases by considering common detection failures. This has two key limitations. First, we require a set of known biases with associated labels within our data to perform the specified computations. 
This is not overly restrictive as many of the explanatory attributes we consider are implicitly defined by standard ground truth labels for detection (\textit{i.e.,} object size, occlusion) or are commonly collected as meta data (\textit{i.e.,} time of day). However, undoubtedly defining more attributes and possible factors of variance will lead to more explanation of observed bias. Our framework can be used with any number of explanatory attributes and is not limited by the 5 we illustrate in this work. 
The second key limitation is that we can not guarantee that our framework reveals \textit{all} possible sources of bias. This will be limited both by our explanatory attribute set as well as the fact that we currently only consider one explanatory attribute at a time. Future expansions of our approach may consider combinations of attributes. The dataset we study, BDD100K, did not have sufficient samples across combinations of explanatory attributes to compute meaningful AP values. 
Finally, we note that we discover correlations and do not claim that there exists a definitive casual relationship between a chosen explanatory attribute and a sensitive attribute.

\noindent\textbf{Bias Mitigation.} The insights revealed by our framework may be used to guide subsequent mitigation strategies. Say our method discovers an explanatory variable $E$ that explains a large proportion of performance variance across a sensitive attribute $A$. The discovered explanation may imply either: 1) model bias and/or 2) data bias. Consider, for example, that for the sensitive attribute, $A$, of income level, one discovers that explanatory attribute, $E$, of object size explains a significant amount of the performance discrepancy across income levels. This would imply an appropriate intervention would be to leverage a detection model with stronger performance on smaller objects. As a second example, consider an explanatory attribute of time of day. In our experiments we found that the dataset collection had disproportionately more night-time images in the low income regions, possibly related to how the dataset was collected. Rather than being an innate aspect of the scene types in the different geographic regions, this implies a bias in the data collection process. Hence, an effective mitigation strategy would be to collect a more equitably distribution of day and nighttime images across all socioeconomic regions. 
 
\end{document}